\documentclass[11pt]{article}
\usepackage{acl}
\usepackage{times}
\usepackage{latexsym}

\usepackage{microtype}
\usepackage{todonotes}
\usepackage{multirow}
\usepackage{array}
\usepackage{graphicx}
\usepackage{url}
\usepackage{amsmath}

\renewcommand{\vec}[1]{\mathbf{#1}}

\title{Learning Rich Representation of Keyphrases from Text}

\author{Mayank Kulkarni \\
  Bloomberg, USA \\
  \texttt{mkulkarni24@bloomberg.net} \\\And
  Debanjan Mahata\thanks{ *This work was done by the author Debanjan Mahata as an employee of Bloomberg} \\
  Moody's Analytics, USA \\
  \texttt{Debanjan.Mahata@moodys.com} \\\AND
    Ravneet Arora \\
  Bloomberg, USA \\
  \texttt{rarora62@bloomberg.net} \\\And
     Rajarshi Bhowmik \\
  Bloomberg, USA \\
  \texttt{rbhowmik6@bloomberg.net}}

\date{}

\begin{document}
\maketitle
\begin{abstract}
 In this work, we explore how to train task-specific language models aimed towards learning rich representation of keyphrases from text documents. We experiment with different masking strategies for pre-training transformer language models (LMs) in discriminative as well as generative settings. In the discriminative setting, we introduce a new pre-training objective - \textbf{K}eyphrase \textbf{B}oundary \textbf{I}nfilling with \textbf{R}eplacement (\textbf{KBIR}), showing large gains in performance (upto 8.16 points in F1) over SOTA, when the LM pre-trained using KBIR is fine-tuned for the task of keyphrase extraction. In the generative setting, we introduce a new pre-training setup for BART - \textbf{KeyBART}, that reproduces the keyphrases related to the input text in the CatSeq format, instead of the denoised original input. This also led to gains in performance (upto 4.33 points in F1@M) over SOTA for keyphrase generation. Additionally, we also fine-tune the pre-trained language models on named entity recognition (NER), question answering (QA), relation extraction (RE), abstractive summarization and achieve comparable performance with that of the SOTA, showing that learning rich representation of keyphrases is indeed beneficial for many other fundamental NLP tasks.

\end{abstract}

\section{Introduction and Background}
\label{sec:intro}




Keyphrases capture the most salient topics of a document and facilitates extreme summarization. Identifying them in an automated way from a text document can be useful for several downstream tasks - classification \cite{hulth2006study}, clustering \cite{hammouda2005corephrase}, summarization \cite{qazvinian2010citation, zhang2004world}, reviewer and document recommendation \cite{augenstein2017semeval}, and many different information retrieval tasks such as enabling semantic and faceted search \cite{sanyal2019enhancing, gutwin1999improving}, query expansion \cite{song2006keyphrase}, and interactive document retrieval \cite{jones1999phrasier}.

Keyphrases could either be \textit{extractive} (part of the document) or \textit{abstractive} (not part of the document). Prior works have referred to them as present and absent keyphrases, respectively. Automatically identifying them entails the process of detecting the extractive \cite{hasan2014automatic} and generating the abstractive keyphrases \cite{ccano2019keyphrasesurvey} from a given document. While extractive approaches have mostly dominated over the generative ones with higher accuracies \cite{ccano2019keyphrase}, the task is far from solved and the performances of the present systems are worse in comparison to many other NLP tasks \cite{liu2010automatic}. Some of the major challenges are the varied length of the documents to be processed, their structural inconsistency and developing strategies that can perform well in different domains.

Most of the prior work on identifying keyphrases using deep learning techniques have concentrated on developing new architectures and frameworks based on different training paradigms such as seq2seq \cite{meng-etal-2017-deep, yuan2018one, zhang2017deep, chen2018keyphrase, ye2018semi, chen2019guided, ye2021one2set}, sequence tagging \cite{alzaidy2019bi}, reinforcement learning \cite{Chan2019NeuralKG}, adversarial training \cite{swaminathan2020preliminary} and game theory \cite{saxena2020keygames}. Although, there has been tremendous progress in learning better semantic and syntactic representation of language at different levels - characters, words, phrases, sentences and documents \cite{liu2020representation}, there hasn't been any effort in learning rich pre-trained representations of keyphrases, which is the major focus of this work.

Transformer language models when pre-trained on large corpora with different pre-training objectives \cite{qiu2020pre} have shown great success in various downstream tasks on fine-tuning, including the tasks of keyphrase extraction \cite{sahrawat2019keyphrase, martinc2020tnt, santosh2020sasake} and generation \cite{liu2020keyphrase}. However, pre-training objectives tailored towards learning better representation of keyphrases that can result in improving the performance of identifying and generating keyphrases from text have not yet been explored. This motivated us to look into this specific problem and make an attempt to answer the following questions:

\noindent \textit{\textbf{Q1}} - \textit{Can we formulate a pre-training objective for language models that can learn better representation of keyphrases?}

Previous work explored training language models for learning better representation of text spans \cite{joshi2020spanbert}, summary sentences \cite{zhang2020pegasus}, and tokens for named entity recognition \cite{yamada2020luke}. To effectively learn rich representation of keyphrases in a BERT like discriminative setup, we propose a new pre-training objective - \textbf{Keyphrase Boundary Infilling with Replacement (KBIR)} (Section \ref{subsec:kbir}) which utilizes a multi-task learning setup for optimizing a combined loss of random token Masked Language Modeling (MLM) \cite{devlin2018bert}, \textbf{Keyphrase Boundary Infilling (KBI)} (Section \ref{subsec:kbi}) and \textbf{Keyphrase Replacement Classification (KRC)} (Section \ref{subsec:krc}).

We also propose a new setup for pre-training BART \cite{lewis2019-bart} - \textbf{KeyBART} (Section \ref{subsec:keybart}), focused towards learning better representation of keyphrases in a generative setting. Instead of reproducing the denoised input text as proposed in the original setup, we produce the keyphrases associated with the input document in the CatSeq \cite{meng-etal-2017-deep} format from a corrupted input.

\noindent \textit{\textbf{Q2}} - \textit{Does learning rich representation of keyphrases in a language model lead to performance gains for the tasks of keyphrase extraction and generation?}

 One of the key contributions of this work is the introduction of KBIR, which is the combination of the KBI and KRC objectives with MLM that helps to learn good representation of keyphrases. This is validated by obtaining SOTA performance for the task of keyphrase extraction on three benchmark datasets (Section \ref{subsec:ke}), surpassing the existing SOTA \cite{duan21_interspeech} by at most 8.16 F1 points on the SemEval 2017 corpus \cite{augenstein2017semeval}.



 We also evaluated the KeyBART approach across five benchmark datasets for the task of keyphrase generation and obtained SOTA performances for both present and absent keyphrases (Section \ref{subsec:kg}). Our best model surpassed the SOTA ONE2SEQ model \cite{ye2021one2set} by 4.33 F1@M points and 0.72 F1@M points on Inspec \cite{Hulth:2003:IAK:1119355.1119383} for present and absent keyphrases respectively.

\noindent \textit{\textbf{Q3}} - \textit{Do rich keyphrase representations aid other fundamental tasks in NLP such as NER, QA, RE and summarization?}

It is to be noted, that although we trained our models on a large corpus of 23 million scientific articles, we find that it performs reasonably well when fine-tuned on datasets that do not belong to the scientific domain for different NLP tasks as shown in Section \ref{subsec:ner}, \ref{subsec:re}, \ref{subsec:qa} and \ref{subsec:summ}. This also suggests that identifying keyphrases in the context of an input text is a fundamental NLP task and a language model trained to learn optimal representation of keyphrases can aid many other tasks. 

To summarize the main contributions that we make in this work are:
\begin{itemize}
    \item We make the first attempt to train task-specific language models in discriminative as well as generative settings geared towards learning rich representation of keyphrases from text.
    \item We introduce a novel pre-training objective \textbf{Keyphrase Boundary Infilling with Replacement (KBIR)} and train a new language model that achieves SOTA performance for the task of keyphrase extraction.
    \item We propose a new setup - \textbf{KeyBART} for pre-training a generative language model for learning better representation of keyphrases and achieve SOTA performance on the task of keyphrase generation.
    \item We also empirically show how learning rich keyphrase representations from text is also useful for other NLP tasks like NER, RE, QA and summarization by achieving near SOTA performances in all of them using our language models trained using KBIR objective and KeyBART settings.
\end{itemize}

We have made our models\footnote{\url{https://huggingface.co/bloomberg/KeyBART}} \footnote{\url{https://huggingface.co/bloomberg/KBIR}} publicly available. We also make our pre-training code\footnote{\url{https://github.com/bloomberg/kbir_keybart}} available. Next, we give a detailed description of the methods that we propose in this work.

\section{Keyphrase Boundary Infilling with Replacement (KBIR)}
\label{subsec:kbir}
\begin{figure*}[t!]
    \centering
    \includegraphics[width=\linewidth]{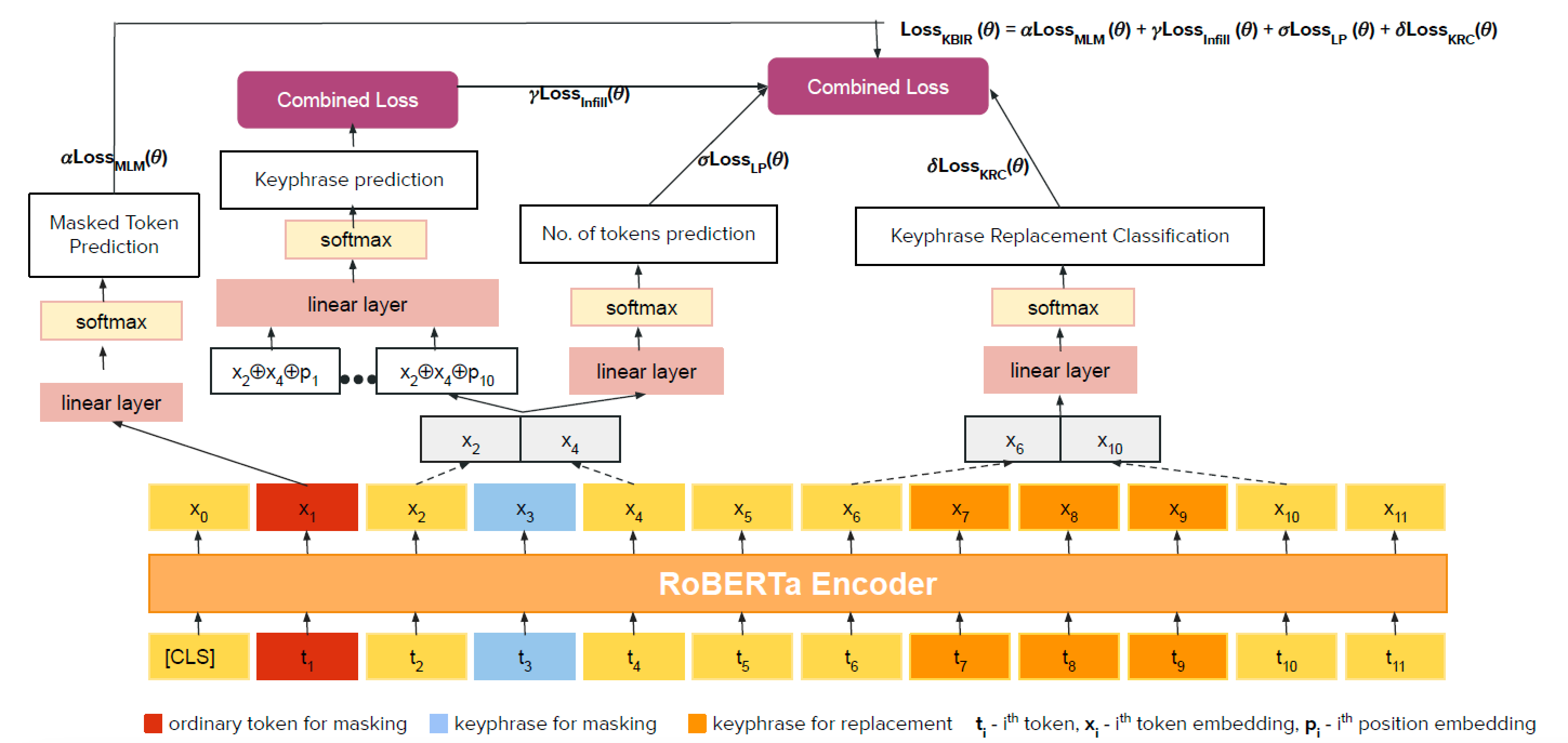}
    \caption{The KBIR model architecture for the training phase. The Random Token MASK is denoted in \textcolor{red}{red}, the Keyphrase MASK is denoted in \textcolor{blue}{blue} and the Replaced Keyphrase in \textcolor{orange}{orange}.}
    \label{fig:keyphrase-rep}
\end{figure*}

In the previous section, we mentioned various methods that aim at learning representations of text spans. Unlike LMs like SpanBERT \cite{joshi2020spanbert} and PEGASUS \cite{zhang2020pegasus} whose primary objective is to learn representations of random or heuristically chosen spans of text, the intuition behind learning good keyphrase representation is to provide the LM the ability to learn spans as well as to identify important phrases (in this case keyphrases) in the context of an input text. This motivated us to devise a framework that can optimize both of these objectives. Towards this effort, we propose a new pre-training objective \textit{Keyphrase Boundary Infilling with Replacement} (KBIR) which is composed of two individual tasks - \textit{Keyphrase Boundary Infilling} (KBI) and \textit{Keyphrase Replacement Classification} (KRC) jointly learnt in a multi-task learning setup as shown in Figure \ref{fig:keyphrase-rep}. We build our framework on top of RoBERTa which implements random token Masked Language Modeling (MLM), therefore making our LM essentially optimizing MLM along with KBI and KRC objectives. In the following section, we describe the individual components of our framework.

\subsection{Keyphrase Boundary Infilling (KBI)}
\label{subsec:kbi}

To effectively learn span representations of keyphrases, we propose a new pre-training objective that builds upon the Span Boundary Objective (SBO) from SpanBERT \cite{joshi2020spanbert} and the Text Infilling setup from BART \cite{lewis2019-bart}. Similar to BART, we replace the entire span, in this case a keyphrase, with a single [MASK] token as shown in Figure \ref{fig:pretraining-strategies} and predict the original tokens using positional embeddings in conjunction with boundary tokens. Text Infilling is a more challenging task than SpanBERT's objective of individual masked token predictions as the model must predict how many tokens correspond to a span \cite{lewis2019-bart}. Different from SpanBERT, which does not penalize incorrect predictions of a sequence of tokens within a masked span, we propose a cumulative loss (Equation \ref{eq:infill}) across all tokens in the masked span to capture intra-span token relationships to learn better span representations. \textit{Text infilling, to the best of our knowledge, has not been explored in a discriminative setup as done in this work}. 

We denote the output of the transformer encoder for each token ${x_l}$ in the sequence $x_{1}, \dots , x_{L}$ as $\vec{x_l}$. However, since the entire span of tokens $(x_{s}, . . . , x_{e})$ of a keyphrase $y_{m}$ is masked with a mask token $x_m$, it is represented with a single vector $\vec{x_m}$, where $(s, e)$ indicates its start and end positions and $m$ represents the index of a masked keyphrase span. We set a maximum possible number of tokens corresponding to a keyphrase span, $T_{max}$ such that $i \in [1, T_{max}]$. We then predict the sequence of tokens to replace $x_{m}$ using the output encodings of the external boundary tokens $x_{s-1}$ and $x_{e+1}$, as well as the position embedding $\vec{p_i}$ of the target token as shown in Equation \ref{eq:infill-y}.

\begin{equation}
\label{eq:infill-y}
\vec{y_{i}} = f(\vec{x_{s-1}}, \vec{x_{e+1}}, \vec{p_i})
\end{equation}

\noindent where positional embeddings use relative positions of the masked tokens with respect to the left boundary token $x_{s-1}$. We use Layer Normalization \cite{ba2016-layer-norm} and GeLU \cite{gelu-hendrycks2016} activation function to represent $f(\Delta)$. We then use the vector representation $\vec{y_i}$ to predict the potential token $x_{i}$ and compute the cumulative cross-entropy loss for each $i$ present within the unmasked $x_{m}$ as shown in Equation \ref{eq:infill}.
\begin{equation}
\label{eq:infill}
\mathcal{L}_{\mathrm{Infill}}(\boldsymbol{\theta})=\sum_{i=1}^{T_{max}} \log p\left(x_{i} | \vec{y_{i}}\right)
\end{equation}

In addition to predicting the actual tokens, we use a classification head to predict the expected number of tokens corresponding to the [MASK] in anticipation of providing a stronger learning signal. Each possible length of the [MASK] is represented as a class and therefore, the number of such classes is equal to the maximum number of possible tokens ($T_{max}$). The architecture used for classifying the number of tokens is a single linear layer which is trained with cross-entropy loss $\mathcal{L}_{\mathrm{LP}}(x_{m}, z_{m})$ along with the infilled masked token $x_{m}$ and the corresponding actual length of the span class $z_{m}$. 

The Keyphrase Boundary Infilling (KBI) objective is formally represented as: 
\begin{equation}
\label{eq:sbi}
\mathcal{L}_{\mathrm{KBI}}(\boldsymbol{\theta})= \alpha \mathcal{L}_{\mathrm{MLM}}(\boldsymbol{\theta}) + \gamma \mathcal{L}_{\mathrm{Infill}}(\boldsymbol{\theta}) + \sigma\mathcal{L}_{\mathrm{LP}}(\boldsymbol{\theta})  
\end{equation}

where $\alpha$, $\gamma$ and $\sigma$ are co-efficients applied to each loss and are primarily used to normalize the losses across the tasks.

\textit{We propose this pre-training objective to be used with keyphrases, however the objective is generic enough to be applied to any spans of text, these could be keyphrases, entities or even random spans.}

\subsection{Keyphrase Replacement Classification (KRC)}
\label{subsec:krc}
Apart from learning representations of keyphrase spans, we wanted our framework to have the ability to identify them within the context of a text input. Motivated by WKLM \cite{xiongetal2019-wklm} that explores pretraining a language model through weak supervision by replacing entities with random entities of the same type that belongs to a knowledge base, we adapt it to replace keyphrases by randomly choosing another keyphrase of variable length from the universe of keyphrases identified in a tagged corpus. The KRC task is then modeled as a binary classification task to determine whether a keyphrase is replaced or retained. 


To implement this strategy, we construct a keyphrase universe by identifying the set of unique keyphrases tagged across the entire dataset. We then randomly shuffle this keyphrase universe and restrict it to 500,000 keyphrases for computational complexity.  We use the concatenated representation of boundary tokens of a keyphrase $x_{s-1}$ and $x_{e+1}$ as input to a linear classifier as shown in Figure \ref{fig:keyphrase-rep}. Given the label $y_{k}$ representing whether a keyphrase was replaced or not, the objective here is to minimize the binary cross-entropy loss $\mathcal{L}_{\mathrm{KRC}}((x_{s-1} + x_{e+1}) , y_{k})$.

Finally, in order to train a LM with an objective of learning good keyphrase representations we use the KBIR pre-training strategy in which we jointly optimize the KBI loss and the KRC loss along with the already existing MLM loss ($\mathcal{L}_{\mathrm{MLM}}(\boldsymbol{\theta})$) in RoBERTa. This is formally shown in Equation \ref{eq:kbir}.

\begin{align}
\label{eq:kbir}
\mathcal{L}_{\mathrm{KBIR}}(\boldsymbol{\theta}) &= 
\alpha \mathcal{L}_{\mathrm{MLM}}(\boldsymbol{\theta}) +
\gamma \mathcal{L}_{\mathrm{Infill}}(\boldsymbol{\theta}) + \nonumber \\
& \qquad \sigma\mathcal{L}_{\mathrm{LP}}(\boldsymbol{\theta}) +
\delta\mathcal{L}_{\mathrm{KRC}}(\boldsymbol{\theta}) 
\end{align}

\section{KeyBART}
\label{subsec:keybart}
We also explored learning a generative LM for the text generation tasks such as keyphrase generation and abstractive summarization. Our hypothesis behind the proposed setup is that masking and replacing task-specific spans, in this case keyphrases, that need to be re-generated should allow the generative model to learn a better representation of surrounding tokens and also the spans themselves.

BART \cite{lewis2019-bart} generates sequences of different lengths from the input perturbed with [MASK] tokens along with token addition and deletion. On similar lines, we propose learning rich keyphrase representations by attempting to generate the \textit{Original Present Keyphrases} in the Catseq format as proposed in \cite{meng-etal-2017-deep} from an input perturbed with token masking, keyphrase masking, and keyphrase replacement as shown in Figure \ref{fig:pretraining-strategies}. We call this method \textbf{KeyBART}. We maintain the order of occurrence of the keyphrases in the original document and remove duplicate occurrences. We also use the same method for finding keyphrase replacements as used in KRC (Section \ref{subsec:krc}). We don't explicitly try to model the keyphrase replacement through a replacement classification head, but rely on learning this implicitly as part of the generation task. Similar to BART, we use a reconstruction loss objective during training which is a cross-entropy loss between the output and set of expected keyphrases.

\section{Experiments and Results}
\label{sec:exp}

\begin{table*}[]
\centering
\scalebox{0.79} {
\begin{tabular}{lrrrrrrrrrrrr}
\hline
\textbf{Model} & \textbf{Batch} & \textbf{Steps} & \textbf{Warmup} & \textbf{$\alpha$} & \textbf{$\gamma$} & \textbf{$\sigma$} & \textbf{$\delta$} & \textbf{MLM} & \textbf{KI} & \textbf{KR} & \textbf{MISL} & \textbf{MKR} \\ \hline
RoBERTa-extended &  4 &  130k &  2.5k & 1.0 & 0.0 & 0.0 & 0.0 & 0.15 & 0.0 & 0.0 & - & - \\ 
KBI &  4 &  130k &  2.5k & 1.0 & 0.33 & 1.0 & 0.0  & 0.15 & 0.2 & 0.0 & 10 & - \\ 
KBIR &  2 &  260k &  5k & 1.0 & 0.33 & 1.0 & 2.0  & 0.05 & 0.2 & 0.4 & 10 & 20 \\ 
KeyBART &  4 &  130k &  2.5k & - & - & - & -  & 0.05 & 0.2 & 0.4 & 10 & 20 \\ 
KeyBART-DOC &  2 & 260k &  5k & - & - & - & -  & 0.05 & 0.2 & 0.4 & 10 & 20 \\ \hline
\end{tabular}
}
\caption{Hyperparameters for our pre-training strategies. All models were trained using 8 Tesla V100 GPUs with the Adam \cite{kingma2015-adam} optimizer and a learning rate of 1e-5. Difference in number of steps is to account for changes in batch size while seeing the same number of data points across training regimes. MLM, Keyphrase Infilling (KI) and Keyphrase Replacement (KR) show the probability of this perturbation occurring in the original text. MLM probability is reduced for KBIR in line with \cite{xiongetal2019-wklm}. Maximum Infill Span Length (MISL) and Maximum Keyphrase Replacements (MKR), are based on averages from OAGKX and computational reasons. The coefficients for the loss are used to normalize the magnitude of loss across the different tasks.} 
\label{tab:pretraining_params}
\end{table*}

\subsection{Language Modeling}
\label{subsec:lm}
\noindent \textbf{Dataset \label{subsec:dataset}} - We use the OAGKX \cite{cano-bojar-2020-two} dataset which consists of 23 million scientific documents across multiple domains sampled from the Open Academic Graph with keyphrases tagged by the authors of the articles. The OAGKX contains keyphrases that appear in the abstract and also those which don't appear in the abstract, making it similar to the keyphrase generation setting with present and absent keyphrases. \textit{To the best of our knowledge, we are the first to explore OAGKX dataset for pre-training a large language model}. 

\begin{table}[]
\centering
\scalebox{0.79} {
\begin{tabular}{lrrrrrr}
\hline
\textbf{Parameter} & \textbf{KE} & \textbf{NER} & \textbf{RE} & \textbf{QA} & \textbf{KG} & \textbf{SUM} \\ \hline
Learning Rate & 5e-5 & 1e-5 & 4e-5 & 3e-5 & 5e-5 & 5e-5 \\ 
Batch         & 4    & 8    & 32   & 48   & 32 & 8 \\ 
Epochs        & 100  & 5    & 10   & 2    & 300k & 20k \\ 
GPUs          & 2    & 1    & 2    & 1    & 4 & 2  \\ \hline
\end{tabular}
}
\caption{Hyperparameters for our downstream task evaluations. KG and SUM specifies steps instead of epochs.} 
\label{tab:downstream_params}
\end{table}

\begin{table}[htbp]
\scalebox{0.85} {
\begin{tabular}{lcccc}
\hline
\textbf{Model} & \textbf{Inspec} & \textbf{SE10} & \textbf{SE17} \\ \hline
RoBERTa+BiLSTM-CRF         & 59.5 & 27.8 & 50.8  \\ 
RoBERTa+TG-CRF             & 60.4 & 29.7 & 52.1 \\ 
SciBERT+Hypernet-CRF       & 62.1 & 36.7 & 54.4 \\ 
RoBERTa+Hypernet-CRF       & 62.3 & 34.8 & 53.3 \\ \hline
RoBERTa-extended-CRF*      & 62.09 & 40.61 & 52.32  \\ 
KBI-CRF*                   & 62.61 & \textbf{40.81} & 59.7\\ 
KBIR-CRF*               & \textbf{62.72} & 40.15 & \textbf{62.56}\\ \hline
\end{tabular}
}
\caption{F1 scores for Keyphrase Extraction on Inspec, SE10 and SE17 datasets (* LMs trained by us).}
\label{tab:ke_results}
\end{table}

During LM pre-training we restricted the length of the input text for each sample to 512 tokens. 

\textit{Note that we do not explicitly tag the keyphrases and use the readily available author tagged keyphrases associated with each document, which is a common practice in the scientific domain.} This setup is analogous to how the Wikipedia corpus is used to perform entity specific pre-training in LUKE \cite{yamada2020luke} and WKLM \cite{xiongetal2019-wklm} among others.

We conducted a preliminary study of using TextRank, a baseline unsupervised keyphrase tagging techniques, to create a weakly supervised dataset but observed only marginal gains. More details are provided in Section \ref{subsec:failed-attempts} that would serve as a potential direction for future work. Additionally, we address Limitations and Ethical Concerns in Appendix \ref{subsec:limitations-ethics}.



\noindent \textbf{Pre-training Strategies} - We train LMs in different settings with different hyperparameters as shown in Table \ref{tab:pretraining_params} (refer Appendix- \ref{subsec:pre-training-strategies} for more details). We pre-train in a discriminative setting with the \textbf{KBIR} method using RoBERTa \cite{liu2019-roberta} pre-trained weights. We also pre-train in a generative setting with the \textbf{KeyBART} method using BART \cite{lewis2019-bart} pre-trained weights. 

\noindent \textbf{Ablations} - Considering computational costs and environmental impact, we conduct a limited set of ablation studies to demonstrate the effectiveness of our proposed methods and also to demonstrate that the gains in performance are not due to additional data.\footnote{We attempted whole word masking keyphrases for both SBO and MLM for BASE model pre-training and observed no significant gains.} We pre-train using basic random token masking strategy as \textbf{RoBERTa-extended} ablating both KBI and KRC from KBIR, using RoBERTa pre-trained weights. We also pre-train using the \textbf{KBI} method, ablating KRC from KBIR using RoBERTa pre-trained weights. We pre-train BART's original denoising autoencoder strategy to recreate the original document as \textbf{KeyBART-DOC} by using BART pre-trained weights.

\subsection{Downstream Task Evaluation}
All our downstream evaluations are performed using HuggingFace Transformer's \cite{wolf2020transformers} RoBERTa or BART architectures to facilitate reproducibility. We also specify all hyperparameters in Table \ref{tab:downstream_params}. We add no additional parameters over RoBERTa or BART for the corresponding downstream evaluation architecture, demonstrating the effectiveness of our updated pre-trained weights.

\subsubsection{Keyphrase Extraction}
\label{subsec:ke}

\begin{table*}[]
\centering
\scalebox{0.76}{
\begin{tabular}{l|cc|cc|cc|cc|cc}
\hline
    & \multicolumn{2}{c|}{\textbf{Inspec}} & \multicolumn{2}{c|}{\textbf{NUS}} & \multicolumn{2}{c|}{\textbf{Krapivin}} & \multicolumn{2}{c|}{\textbf{SemEval}} & \multicolumn{2}{c}{\textbf{KP20k}} \\ 
\textbf{Model} & \textbf{F1@5} & \textbf{F1@M} & \textbf{F1@5} & \textbf{F1@M} & \textbf{F1@5} & \textbf{F1@M} & \textbf{F1@5} & \textbf{F1@M} & \textbf{F1@5} & \textbf{F1@M} \\ \hline
catSeq \cite{yuan2018one} & 22.5 & 26.2 & 32.3 & 39.7 & 26.9 & 35.4 & 24.2 & 28.3 & 29.1 & 36.7\\ 
catSeqTG \cite{chen2019guided} & 22.9 & 27 & 32.5 & 39.3 & 28.2 & 36.6 & 24.6 & 29.0 & 29.2 & 36.6\\ 
catSeqTG-2RF1 \cite{Chan2019NeuralKG} & 25.3 & 30.1 & \textbf{37.5} & 43.3 & 30 & 36.9 & 28.7 & 32.9 & 32.1 & 38.6\\ 
GANMR \cite{swaminathan2020preliminary} & 25.8 & 29.9 & 34.8 & 41.7 & 28.8 & 36.9 & - & - & 30.3 & 37.8\\ 
ExHiRD-h \cite{chen-etal-2020-exclusive} & 25.3 & 29.1 & - & - & 28.6 & 34.7 & 28.4 & 33.5 & 31.1 & 37.4\\ 
Transformer \cite{ye2021one2set} & 28.15 & 32.56 & 37.07 & 41.91 & \textbf{31.58} & 36.55 & \textbf{28.71} & 32.52 & \textbf{33.21} & 37.71\\ \hline
BART* &  23.59 & 28.46 & \textbf{\textit{35.00}} & 42.65 & 26.91 & 35.37 & 26.72 & 31.91 & 29.25 & 37.51\\  
KeyBART-DOC* & 24.42 & 29.57 & 31.37 & 39.24 & 24.21 & 32.60 & 24.69 & 30.50 & 28.82 & 37.59\\ 
KeyBART* & 24.49 & 29.69 & 34.77 & \textbf{43.57} & \textbf{\textit{29.24}} & \textbf{38.62} & \textbf{\textit{27.47}} & \textbf{33.54} & \textbf{\textit{30.71}} & \textbf{39.76} \\
KeyBART* (no finetune) & \textbf{30.72} & \textbf{36.89} & 18.86 & 21.67 & 18.35 & 20.46 & 20.25 & 25.82 & 12.57 & 15.41 \\
\hline
\end{tabular}
}
\caption{Keyphrase Generation for Present Keyphrases. SOTA is marked in \textbf{Bold} and our best performing models as \textbf{\textit{Bold-Italicized}}.}
\label{tab:KG_Present_results}
\end{table*}

\begin{table*}[]
\centering
\scalebox{0.76}{
\begin{tabular}{l|cc|cc|cc|cc|cc}
\hline
    & \multicolumn{2}{c|}{\textbf{Inspec}} & \multicolumn{2}{c|}{\textbf{NUS}} & \multicolumn{2}{c|}{\textbf{Krapivin}} & \multicolumn{2}{c|}{\textbf{SemEval}} & \multicolumn{2}{c}{\textbf{KP20k}} \\ 
\textbf{Model} & \textbf{F1@5} & \textbf{F1@M} & \textbf{F1@5} & \textbf{F1@M} & \textbf{F1@5} & \textbf{F1@M} & \textbf{F1@5} & \textbf{F1@M} & \textbf{F1@5} & \textbf{F1@M} \\ \hline
catSeq \cite{yuan2018one} & 0.4 & 0.8 & 1.6 & 2.8 & 1.8 & 3.6 & 1.6 & 2.8 & 1.5 & 3.2 \\ 
catSeqTG \cite{chen2019guided} & 0.5 & 1.1 & 1.1 & 1.8 & 1.8 & 3.4 & 1.1 & 1.8 & 1.5 & 3.2 \\ 
catSeqTG-2RF1 \cite{Chan2019NeuralKG} & 1.2 & 2.1 & 1.9 & 3.1 & 3.0 & 5.3 & \textbf{2.1} & \textbf{3.0} & 2.7 & \textbf{5.0} \\ 
GANMR \cite{swaminathan2020preliminary} & 1.3 & 1.9 & 2.6 & 3.8 & \textbf{4.2} & 5.7 & - & - & \textbf{3.2} & 4.5 \\ 
ExHiRD-h \cite{chen-etal-2020-exclusive} & 1.1 & 2.2 & - & - & 2.2 & 4.3 & 1.7 & 2.5 & 1.6 & 3.2 \\ 
Transformer \cite{ye2021one2set} & 1.02 & 1.94 & \textbf{2.82} & \textbf{4.82} & 3.21 & 6.04 & 2.05 & 2.33 & 2.31 & 4.61 \\ \hline
BART* & 1.08 & 1.96 & \textbf{\textit{1.80}} & \textbf{\textit{2.75}} & 2.59 & 4.91 & 1.34 & 1.75 & 1.77 & 3.56\\ 
KeyBART-DOC* & 0.99 & 2.03 & 1.39 & 2.74 & 2.40 & 4.58 & 1.07 & 1.39 & 1.69 & 3.38\\   
KeyBART* &  0.95 & 1.81 & 1.23 & 1.90 & \textbf{\textit{3.09}} & \textbf{6.08} & \textbf{\textit{1.96}} & \textbf{\textit{2.65}} & \textbf{\textit{2.03}} & \textbf{\textit{4.26}} \\
KeyBART* (no finetune) & \textbf{1.83} & \textbf{2.92} & 1.46 & 2.19 & 1.29 & 2.09 & 1.12 & 1.45 & 0.70 & 1.14 \\ \hline
\end{tabular}
}
\caption{Keyphrase Generation for Absent Keyphrases. SOTA is marked in \textbf{Bold} and our best performing models as \textbf{\textit{Bold-Italicized}}.}
\label{tab:KG_Absent_results}
\end{table*}

\begin{table}[]
\centering
\scalebox{0.76} {
\begin{tabular}{lr}
\hline
\textbf{Model} & \textbf{F1}\\ \hline
LSTM-CRF \cite{lample-etal-2016-neural} &  91.0 \\ 
ELMo \cite{peters-etal-2018-deep}  & 92.2 \\ 
BERT \cite{devlin2018bert}  & 92.8 \\ 
\cite{akbik-etal-2019-pooled} &  93.1 \\ 
\cite{baevski-etal-2019-cloze} &  93.5 \\ 
LUKE \cite{yamada2020luke} &  \textbf{94.3} \\ 
LUKE w/o entity attention &  94.1 \\ \hline
RoBERTa \cite{yamada2020luke}   & 92.4  \\ 
RoBERTa-extended*   & 92.54   \\ 
KBI*       & 92.73   \\ 
KBIR*    & \textbf{\textit{92.97}}  \\ \hline
\end{tabular}
}
\caption{Named Entity Recognition results on  CONLL-2003. SOTA is marked in \textbf{Bold} and our best performing models as \textbf{\textit{Bold-Italicized}}.}
\label{tab:ner_results}
\end{table}

We report performance of our models for Keyphrase Extraction (KE) on Inspec \cite{inspec}, SemEval-2010 (SE10) \cite{kim-etal-2010-semeval}, and SemEval-2017 (SE17) \cite{augenstein2017semeval}. \cite{sahrawat2019keyphrase} explored KE as a sequence tagging task with contextual embeddings and demonstrate the effectiveness of a CRF. We compare our performance with RoBERTa+BiLSTM-CRF \cite{sahrawat2019keyphrase}, RoBERTa+TG-CRF \cite{chen2019guided}, previous state-of-the-art model RoBERTa+Hypertnet-CRF, and SciBERT+Hypernet-CRF \cite{duan21_interspeech}. However, different from these architectures, we do not use a LSTM/BiLSTM layer between the contextualized embeddings and the CRF. We fine-tune all the pre-trained language models on B-I-O tagged datasets for KE\footnote{https://github.com/midas-research/keyphrase-extraction-as-sequence-labeling-data}. We use hyperparameters specified in \cite{sahrawat2019keyphrase} and F1-score is used as the evaluation metric.

Table \ref{tab:ke_results} shows our pre-trained LMs outperform SOTA by significant margins across all three datasets despite having fewer parameters. While RoBERTa-extended, shows gains over RoBERTa+BiLSTM-CRF, this is expected since the domain of the continued pre-training data is more in line for KE evaluation. However, the models that explicitly learn keyphrase representations such as KBI and KBIR significantly outperform RoBERTa-extended. We believe the slight gain for SemEval-2010 is because of the small size of the dataset (130 - train, 100 - test).

\subsubsection{Keyphrase Generation}
\label{subsec:kg}

We evaluate keyphrase generation (KG) performance on Inspec \cite{inspec}, NUS \cite{NUSDataset}, Krapivin \cite{krapivin2009large}, SemEval \cite{kim-etal-2010-semeval} and KP20K \cite{meng-etal-2017-deep}. The task is to generate the \textit{CatSeq} output of the present and absent keyphrases for a given concatenated title and abstract, as done in previous works \cite{meng-etal-2017-deep, chen2019guided, yuan2018one}. We use the \textit{PresAbs} ordering of the keyphrases as that was shown to be the most effective representation in \cite{meng-etal-2021-empirical}. Further, we only train a single model by fine-tuning on the KP20K dataset and perform inference on all the test datasets. Similar to \cite{meng-etal-2021-empirical} we use a beam width of 50 for beam search and restrict our maximum generated sequence length to 40 tokens. For our evaluation we use macro-averaged F1@5 and F1@M as in \cite{Chan2019NeuralKG} and \cite{chen-etal-2020-exclusive} for both present and absent keyphrase generation. F1@M evaluates all the keyphrases predicted by the model with the ground-truth keyphrases. F1@5, as the name suggests evaluates only the first 5 keyphrases, however when there are fewer than five keyphrases, random incorrect keyphrases are appended till it reaches five predictions. \cite{Chan2019NeuralKG} show that without this appending F1@M is the same as F1@5, when predictions are fewer than five. \cite{ye2021one2set} also present a ONE2SET training paradigm and for a fair comparison we compare to their Transformer (ONE2SEQ) results, since we also train in the ONE2SEQ paradigm and not ONE2SET.

In Table \ref{tab:KG_Present_results} and Table \ref{tab:KG_Absent_results} we see that KeyBART is the most effective pre-training method achieving SOTA on most datasets for F1@M in present and absent KG. We believe our choice of perturbation of the input during the pre-training setup makes this model robust and helps it identify and generate keyphrases more effectively. We also observe that our results for F1@5 aren't as competitive as F1@M and we believe this is because our model tends to favor predicting fewer than 5 keyphrases and thus tends to suffer from the random addition of keyphrases for F1@5. More concretely, the average predicted keyphrases per document for SemEval is 2.51, NUS is 2.86, Krapivin is 2.86, Inspec is 3.09 and KP20k is 2.73. The Inspec dataset is anomalous where the non-finetuned model performs significantly better, demonstrating the effectiveness of the KeyBART training strategy.


\subsubsection{Named Entity Recognition}
\label{subsec:ner}
We report the performance of different models for the task of NER by conducting experiments on CoNLL-2003 dataset \cite{sang2003introduction}. 


Table \ref{tab:ner_results} demonstrate that KBI and KBIR have performance gains over RoBERTa on CoNLL-2003. With RoBERTa-extended, we see that only continued pre-training with the MLM objective results in minor gains. However, when we inspect the results for KBI and KBIR, we see consistent jumps in performance showing how both these architectures contribute in learning richer representations that directly impact NER performance. We hypothesize that KBIR is more effective at NER than KBI because the additional keyphrase replacement classification task builds richer boundary token representations making entity identification potentially easier. The results are also fairly competitive with SOTA NER models in literature despite the fact that we did not attempt modeling entities explicitly like existing SOTA model \cite{yamada2020luke}.


\subsubsection{Relation Extraction} 
\label{subsec:re}
The relation extraction (RE) task predicts relations among pairs of entity mentions in a text. We fine-tuned our models for the sentence-level relation extraction task using the popular TACRED benchmark dataset \cite{zhang-etal-2017-position}. TACRED contains more than 100,000 sentences with entities that belong to 23 different fine-grained semantic types and with 42 different relations among entities. To fine-tune our models, we modified the input sequences to mark the start and end of the subject entity with @ and the object entity with \#. We use the final layer representation of the [CLS] token as the input to a multi-class classifier. 

The results in the top half of Table~\ref{tab:re_results} are reported from the respective papers that use various input formatting strategy. Similar to \cite{zhou2021improved}, we also observe that a model's performance depends heavily on the formatting of the input sequence. All models in the bottom half of the table are trained with the same input format mentioned above. We observe that our KBIR model performs slightly worse than the original RoBERTa model. We also observe similar trends for KBI and RoBERTa-extended models. We conjecture that the domain shift of the pre-training corpus is responsible for the slight performance degradation.


\begin{table}[]
\centering
\scalebox{0.76} {
\begin{tabular}{lrr}
\hline
\textbf{Model} & \textbf{EM} & \textbf{F1} \\ \hline
BERT \cite{devlin2018bert} &  84.2 &  91.1 \\ 
XLNet \cite{yang2019-xlnet} &  89.0 &  94.5 \\ 
ALBERT \cite{lan2019-albert} &  89.3 &  94.8 \\ 
LUKE \cite{yamada2020luke} &  \textbf{89.8} &  \textbf{95.0} \\ 
LUKE w/o entity attention &  89.2 & 94.7 \\ \hline
RoBERTa \cite{liu2019-roberta}         & 88.9 & 94.6 \\ 
RoBERTa-extended*   & 88.88 & 94.55    \\ 
KBI*       & 88.97 & 94.7 \\ 
KBIR*    & \textbf{\textit{89.04}} & \textbf{\textit{94.75}}  \\ \hline
\end{tabular}
}
\caption{Question Answering results on SQuAD v1.1 on the DEV set. State-of-the-art is marked in \textbf{Bold} and our best performing models as \textbf{\textit{Bold-Italicized}}.}
\label{tab:qa_results}
\end{table}

\subsubsection{Question Answering} 
\label{subsec:qa}
The relation between question answering (QA) and KE has been explored to some extent in \cite{subramanian-etal-2018-neural}, which leverages keyphrase extraction for question generation. Motivated by their work, we evaluate our models on SQuAD v1.1 \cite{rajpurkar2016squad} dataset for the extractive question answering task. For all the models, we use a maximum sequence length of 512 with a sliding window of size 128. 

Table \ref{tab:qa_results} reports the F1 and Exact Match (EM) scores achieved by different model architectures on the DEV set. We observe improved performance with KBI and KBIR as compared to RoBERTa. We have an interesting observation where RoBERTa-extended performs worse than RoBERTa and we conjecture that it is because of the domain shift in the pre-training data which comprises scientific articles. On the other hand, the models trained with keyphrase pre-training objectives are fairly competitive with the SOTA QA models. We explicitly include LUKE w/o entity attention since that removes the entity-aware attention module, making it slightly more comparable to our setup. We observe that KBIR outperforms it by a slim margin in F1. However, the performance is slightly lower in the EM scores. Note that our model does not yield the similar performance in EM as it does in F1 when compared to SOTA. A potential reason for this is that our model is more likely to identify keyphrases as answers.

\begin{table}[]
\centering
\scalebox{0.76} {
\begin{tabular}{lr}
\hline
\textbf{Model} & \textbf{F1}\\ \hline
BERT \cite{zhang-etal-2019-ernie} & 66.0 \\
C-GCN \cite{zhang-etal-2018-graph} & 66.4 \\ 
ERNIE \cite{zhang-etal-2019-ernie} & 68.0 \\ 
SpanBERT \cite{joshi2020spanbert} & 70.8 \\ 
MTB \cite{baldini-soares-etal-2019-matching} & 71.5 \\ 
KnowBERT \cite{peters-etal-2019-knowledge} &  71.5 \\ 
KEPLER \cite{wang2019-kepler} & 71.7 \\ 
K-Adapter \cite{wang-etal-2021-k}  & 72.0 \\ 
LUKE \cite{yamada2020luke} & \textbf{72.7} \\ 
LUKE w/o entity attention & 72.2 \\ \hline
RoBERTa \cite{wang-etal-2021-k}   & 71.3  \\ 
RoBERTa-extended*  & 70.94   \\ 
KBI*        & 70.71   \\ 
KBIR*    & \textbf{\textit{71.0}}  \\ \hline
\end{tabular}
}
\caption{Relation Extraction results on TACRED. State-of-the-art is marked in \textbf{Bold} and our best performing models as \textbf{\textit{Bold-Italicized}}.}
\label{tab:re_results}
\end{table}


\begin{table}[t]
\centering
\scalebox{0.76}{
\begin{tabular}{lccc}
\hline
\textbf{Model} & \textbf{R1} & \textbf{R2} & \textbf{RL} \\ \hline
BART \cite{lewis2019-bart} & 44.16 & 21.28 & 40.9      \\ \hline
BART*             & 42.93 & 20.12 & 39.72 \\ 
KeyBART-DOC*  & 42.92 & 20.07 & 39.69  \\ 
KeyBART*   & \textbf{\textit{43.10}} & \textbf{\textit{20.26}} & \textbf{\textit{39.90}}  \\ \hline
\end{tabular}
}
\caption{Summarization results on CNN/DailyMail dataset. Our best performing models are marked as \textbf{\textit{Bold-Italicized}}.}
\label{tab:summ_results}
\end{table}

\subsubsection{Summarization}
\label{subsec:summ}
We fine-tune BART \cite{lewis2019-bart}, KeyBART-DOC and KeyBART on the CNN DailyMail \cite{cnn-dm} summarization dataset (SUM). Keyphrase Generation is also considered as an extreme form of summarization and therefore, we expect to see improved performance for the summarization task. Since we were unable to reproduce the original BART scores for R1, R2 and RLSum, we used the reported hyperparameters to reproduce the results to best of our ability, accounting for minor implementation differences in framework versions. We hope this provides a more fair comparison with our model results. We do not claim SOTA for summarization models, rather want to demonstrate that there are potential performance gains by training on a keyphrase specific objective. This is demonstrated in Table \ref{tab:summ_results} where we see that the standard denoising autoencoder setup results in marginal losses. However, training with the keyphrase generation objective improves the ROUGE scores across the board when compared with BART trained on the same dataset.

\section{Qualitative Analysis}
\label{subsec:qualitative_analysis_examples}

We perform a qualitative analysis on the SemEval-2010 dataset as it is the only common dataset between KE and KG tasks by leveraging predictions from the best performing models. We present examples in Table \ref{tab:qualitative_results}, in the Appendix, which captures the ground truth, extracted keyphrases and generated keyphrases (present and absent) for a given document from the SemEval dataset using our best performing models on the respective tasks. We observe that when the model tends to generate more keyphrases, it typically relies on the copy mechanism and hence most of the generated keyphrases are present in the text itself (Example 1). We also observe that when absent keyphrases are generated accompanied by a large number of generated keyphrases, they are usually a combination of two or more words directly present in the text such as `user study' (Example 3). The example discusses how the authors study user behavior, potentially making `user study' a fair prediction, however the ground truth would penalize the model if this was in the training phase.  Finally, we observe more generated keyphrases when the model isn't able to identify keyphrases in text and doesn't rely heavily on the copy mechanism, but on it's understanding of the text. This results in keyphrases such as `natural language processing' (Example 2). Although the prediction is not in the ground-truth, it aligns with the mentions of `question answering' and `linguistics'. This demonstrates that the model is indeed able to generate meaningful absent keyphrases. However, we observe that the model is not able to learn or infer world knowledge required to produce the absent keyphrases in the ground-truth. For keyphrase extraction, we see that the model tends to tag phrases more frequently than previous models, improving recall. We hypothesize that it is due to the model having a better understanding of keyphrases in a document because of the keyphrase masking perturbation and also the KRC task.

\section{Other Experiments}
\label{subsec:failed-attempts}
We also attempted to use a combination of the Wikipedia (English) dump and S2ORC \cite{s2orc-corpus} corpora for pre-training our models. In order to obtain keyphrase tags for data at a large scale, we employed TextRank \cite{mihalcea2004textrank} on each document in the corpora. We set the maximum number of keyphrases to 10 for the TextRank algorithm and considered all keyphrases tagged by TextRank. We created random splits for our dataset to generate a train and development (dev) set. However, we found that keyphrases tagged in this manner added a lot of noise to the dataset and resulted in only marginal overall gains. 

To train a keyphrase specific language model, we also used a combined generative and discriminative approach as introduced in ELECTRA \cite{clark2020electra}. The generative approach made the model predict the masked tokens that are part of a keyphrase. In the discriminative approach, the original sequence was perturbed by replacing the tokens of a keyphrase with another semantically unrelated keyphrase. We were unable to stabilize the training for such a setup and didn't get promising results.

\section{Conclusion and Future Work}
We explored LMs capable of learning rich representations of keyphrases that achieve SOTA performance across multiple datasets for keyphrase extraction and generation tasks. Towards this effort, we proposed a new pre-training objective KBIR and a new training setup KeyBART. The trained LMs demonstrate their effectiveness by achieving SOTA or near SOTA performance for various downstream NLP tasks when fine-tuned on benchmark datasets spanning across multiple domains. As a next step, we would like to probe our LMs to understand them more and also gauge their effectiveness for the tasks of cross-domain keyphrase extraction and generation. We would also like to explore scaling these approaches to more datasets by revisiting more sophisticated unsupervised keyphrase tagging methods.

\clearpage
\bibliography{anthology,acl2020}
\bibliographystyle{acl_natbib}
\clearpage
\section{Appendix}
\subsection{Limitations and Ethical Concerns}
\label{subsec:limitations-ethics}
Experiments were conducted using a private infrastructure, which has a carbon efficiency of 0.432 kgCO$_2$eq/kWh. A cumulative of 6,144 hours of computation was performed on hardware of type Tesla V100-SXM2-32GB (TDP of 300W).We calculate that the combined cost of training all these models is 796.26 KGs of $CO_2$ eq. Estimations were conducted using the \href{https://mlco2.github.io/impact#compute}{Machine Learning Impact calculator} presented in \cite{lacoste2019quantifying}. Given the computational cost and environmental impact we restrict the number of experiment settings and ablation studies conducted in the pre-training stage - so there may be some hyperparameters that could be further optimized but have not been realized and also more robust results to further demonstrate the effectiveness our proposed approach. 

The OAGKX dataset we train on is made publicly available with a Creative Commons License 4.0 and is primarily focused on scientific documents from the Open Academic Graph. This decreases the potential for the pre-trained model to imbibe offensive content and also in not generating the same.  

We also acknowledge that the current setup uses a dataset that already contains pre-tagged keyphrases similar to how Wikipedia leverages entities and our attempt at using a basic unsupervised keyphrase tagging technique did not yield much success as seen in Section \ref{subsec:failed-attempts}. We believe that further research in exploring more sophisticated techniques for unsupervised keyphrase tagging would help overcome this hurdle. Further our proposed approaches should work on entities from Wikipedia or even random spans from the BookCorpus, and we encourage exploration of the same.



\subsection{Pre-training Strategies}
\label{subsec:pre-training-strategies}
\begin{figure*}[t!]
    \centering
    \includegraphics[width=\linewidth]{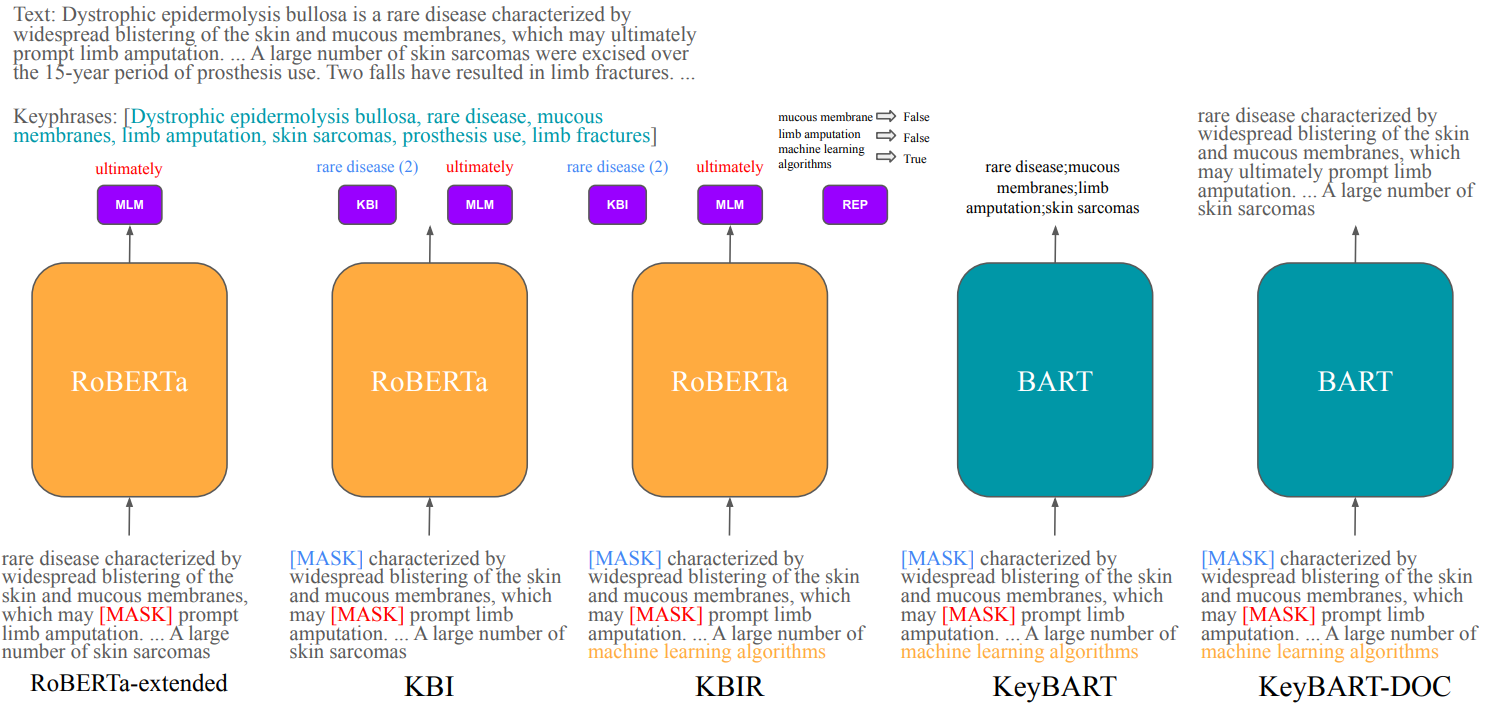}
    \caption{Pre-training Strategies, the keyphrase present in the text are highlighted in \textcolor{teal}{teal} and every perturbation in the form of a random MASK is represented in \textcolor{red}{red}, keyphrase MASK is represented in \textcolor{blue}{blue} and Keyphrase Replacement is represented in \textcolor{orange}{orange}.}
    \label{fig:pretraining-strategies}
\end{figure*}

Figure \ref{fig:pretraining-strategies} provides a visual representation of the various masking strategies we deploy, a more detailed description of each stage is available in the subsections below.

We train LMs in different settings and hyperparameters as listed in Table \ref{tab:pretraining_params}. 

\noindent \textbf{Discriminative Setting} - We pre-train three language models in the discriminative setting as described below. All of them use the pre-trained weights of RoBERTa-large\footnote{https://huggingface.co/roberta-large} as the initial weights and are trained by continuing the learning of the parameters on the OAGKX dataset using our pre-training strategies.

\begin{itemize}
    \item{\textbf{RoBERTa-extended}} -  Previous work \cite{gururangan-etal-2020-dont} has shown that adding more data to pre-training a language model typically results in better downstream performance. To verify that our performance gains stem from modeling improvements and the new pre-training objectives proposed by us rather than addition of data, we extend the training of RoBERTa-large on the OAGKX corpus. We call this model \textit{RoBERTa-extended}. This also ensures fair comparison of the LMs trained by us using our pre-training objectives with that of RoBERTa.
    
    \item{\textbf{KBI}} - During the pre-training of the LM with the KBI objective, we employ both token masking and keyphrase masking strategies as shown in Figure \ref{fig:pretraining-strategies} and explained in Section \ref{subsec:kbi}. We randomly mask 15\% of the tokens that are not included in keyphrase spans. We additionally mask 20\% of the keyphrase spans with a single [MASK] token. We restrict the maximum number of tokens for a keyphrase mask span to 10, based on the average keyphrase length reported in \cite{cano-bojar-2020-two}.
    
    \item{\textbf{KBIR}} - While pre-training the LM with the KBIR objective we employ 5\% token masking, in line with the findings reported in \cite{xiongetal2019-wklm} and 20\% of keyphrases are masked through keyphrase masking, with a maximum possible span size of 10 as in the KBI LM. Additionally, we replace 40\% of the non-masked keyphrases with randomly sampled keyphrases from the keyphrase universe as explained in Section \ref{subsec:krc}. We restrict the maximum number of keyphrases to be replaced to no more than 20, restricted by the computational complexity of the problem. Figure \ref{fig:keyphrase-rep} shows the final architecture, with a multi-task learning objective trained with a weighted combined loss.
\end{itemize}

\noindent \textbf{Generative Setting} - In the generative setting we pre-train two language models as described below. In both the models we continue the training of the weights of BART-large\footnote{https://huggingface.co/facebook/bart-large} on our corpus using our pre-training strategies.
\begin{itemize}
    \item{\textbf{KeyBART}} - We perform token masking, keyphrase masking and keyphrase replacement with same masking hyperparameters as KBIR on the input text and pre-train the model to predict the original keyphrases in Catseq format following the setup explained in Section \ref{subsec:keybart}.
    \item{\textbf{KeyBART-DOC}} - This setup uses the same input denoising settings as KeyBART, with the only difference in the output, where KeyBART generates the keyphrases associated with the document in Catseq format, whereas KeyBART-DOC similar to BART generates the original denoised input.
\end{itemize}

We use the exact same data in all the pre-training setups as explained above. We increase the number of steps while decreasing the batch size such that all the models see the data the same number of times (i.e., 2 epochs). \textit{The batch size is only reduced to accommodate increases in memory usage in the model pre-training}.

\begin{table*}[htbp]
\small
\centering
    \begin{tabular}{p{0.95\linewidth}}
    \hline
    \textbf{Input Text}: On The Complexity of Combinatorial Auctions : Structured Item Graphs and Hypertree Decompositions. The winner determination problem in combinatorial auctions is the problem of determining the allocation of the items among the bidders that maximizes the sum of the accepted bid prices. While this problem is in general NPhard, it is known to be feasible in polynomial time on those instances whose associated item graphs have bounded treewidth called structured item graphs. Formally, an item graph is a graph whose nodes are in one-to-one correspondence with items, and edges are such that for any bid, the items occurring in it induce a connected subgraph. Note that many item graphs might be associated with a given combinatorial auction, depending on the edges selected for guaranteeing the connectedness. In fact, the tractability of determining whether a structured item graph of a fixed treewidth exists and if so, computing one was left as a crucial open problem. In this paper, we solve this problem by proving that the existence of a structured item graph is computationally intractable, even for treewidth 3. Motivated by this bad news, we investigate different kinds of structural requirements that can be used to isolate tractable classes of combinatorial auctions. We show that the notion of hypertree decomposition, a recently introduced measure of hypergraph cyclicity, turns out to be most useful here. Indeed, we show that the winner determination problem is solvable in polynomial time on instances whose bidder interactions can be represented with dual hypergraphs having bounded hypertree width. Even more surprisingly, we show that the class of tractable instances identified by means of our approach properly contains the class of instances having a structured item graph. \\
    \hline
    \textbf{Extracted Keyphrases}: [combinatorial auctions]; [structured item graphs]; [hypertree decompositions]; [item graphs]; [treewidth]; [bidders]; [hose nodes]; [hypertree cyclicity]; [polynomial time] \\
    \textbf{Generated Keyphrases}: [combinatorial auctions]; [structured item graphs]; [treewidth]; [hypergraphs]; [hypertree decompositions] \\
    \textbf{Ground Truth}: [hypergraph]; [structured item graph]; [polynomial time]; [combinatorial auction]; [fixed treewidth]; [accepted bid price]; [hypertree decomposition]; structured item graph complexity; simplification of the primal graph; hypertree based decomposition method; hypergraph hg; the primal graph simplification; well known mechanism for resource and task allocation; complexity of structured item graph;  \\
    \hline
    \hline
    \textbf{Input Text}: Interesting Nuggets and Their Impact on Definitional Question Answering. Current approaches to identifying definitional sentences in the context of Question Answering mainly involve the use of linguistic or syntactic patterns to identify informative nuggets. This is insufficient as they do not address the novelty factor that a definitional nugget must also possess. This paper proposes to address the deficiency by building a Human Interest Model from external knowledge. It is hoped that such a model will allow the computation of human interest in the sentence with respect to the topic. We compare and contrast our model with current definitional question answering models to show that interestingness plays an important factor in definitional question answering. \\
    \hline
    \textbf{Extracted Keyphrases}: [interesting nuggets]; [definitional sentences]; [question answering]; [nuggets]; [novelty]; [definitional nuggets]; [human interest model]; [human interest] \\
    \textbf{Generated Keyphrases}: [definitional question answering]; natural language processing \\
    \textbf{Ground Truth}: [human interest]; [use of linguistic]; [interesting nugget]; [definitional question answer]; [informative nugget]; [interest]; [computation of human interest]; sentence fragment; unique quality; manual labor; news corpus; baseline system; human interest computation; human reader; linguistic use; question topic; external knowledge; surprise factor; lexical pattern \\
    \hline
    \hline
    \textbf{Input Text}: The Influence of Caption Features on Clickthrough Patterns in Web Search. Web search engines present lists of captions, comprising title, snippet, and URL, to help users decide which search results to visit. Understanding the influence of features of these captions on Web search behavior may help validate algorithms and guidelines for their improved generation. In this paper we develop a methodology to use clickthrough logs from a commercial search engine to study user behavior when interacting with search result captions. The findings of our study suggest that relatively simple caption features such as the presence of all terms query terms, the readability of the snippet, and the length of the URL shown in the caption, can significantly influence users ' Web search behavior. \\
    \hline
    \textbf{Extracted Keyphrases}: [influence]; [caption features]; [clickthrough patterns]; [web search]; [snippet]; [methodology]; [clickthrough logs] \\
    \textbf{Generated Keyphrases}: [web search]; [clickthrough]; [captions]; user study \\
    \textbf{Ground Truth}: [clickthrough pattern]; [snippet]; [web search behavior]; [web search]; [caption feature]; summarization; extractive summarization; significant word; query log; human factor; clickthrough inversion; query term match; query re formulation \\
    \hline
    \hline
    \end{tabular}
\caption{Sample keyphrases extracted by KBI-REP-CRF and generated by KeyBART on the SemEval-2010 dataset. Present keyphrases are marked with square brackets.}
\label{tab:qualitative_results}
\end{table*}

\end{document}